\documentclass[sigconf,nonacm]{acmart}
\usepackage[utf8]{inputenc}
\usepackage{graphicx}
\usepackage{amsmath}
\usepackage{hyperref}
\usepackage{natbib}
\usepackage{booktabs}
\usepackage{stfloats}

\title{Stability-Aware Prompt Optimization \\for Clinical Data Abstraction}

\begin{document}

\author{Arinbjörn Kolbeinsson}
\affiliation{
\institution{Century Health}
% \city{City}
\country{}
}
\email{arinbjorn.kolbeinsson@century.health}

\author{Daniel Timbie}
\affiliation{
\institution{Century Health}
% \city{City}
\country{}
}
\email{daniel.timbie@century.health}

\author{Sajjan Narsinghani}
\affiliation{
\institution{Century Health}
% \city{City}
\country{}
}
\email{sajjan.narsinghani@century.health}

\author{Sanjay Hariharan}
\affiliation{
\institution{Century Health}
% \city{City}
\country{}
}
\email{sanjay.hariharan@century.health}

\begin{abstract}
Large language models used for clinical abstraction are sensitive to prompt wording, yet most work treats prompts as fixed and studies uncertainty in isolation. We argue these should be treated jointly. Across two clinical tasks (MedAlign applicability/correctness and MS subtype abstraction) and multiple open and proprietary models, we measure prompt sensitivity via flip rates and relate it to calibration and selective prediction. We find that higher accuracy does not guarantee prompt stability, and that models can appear well-calibrated yet remain fragile to paraphrases. We propose a dual-objective prompt optimization loop that jointly targets accuracy and stability, showing that explicitly including a stability term reduces flip rates across tasks and models, sometimes at modest accuracy cost. Our results suggest prompt sensitivity should be an explicit objective when validating clinical LLM systems.
\end{abstract}

\maketitle

\section{Introduction}

\begin{figure*}[b!]
\centering
\includegraphics[width=0.85\textwidth]{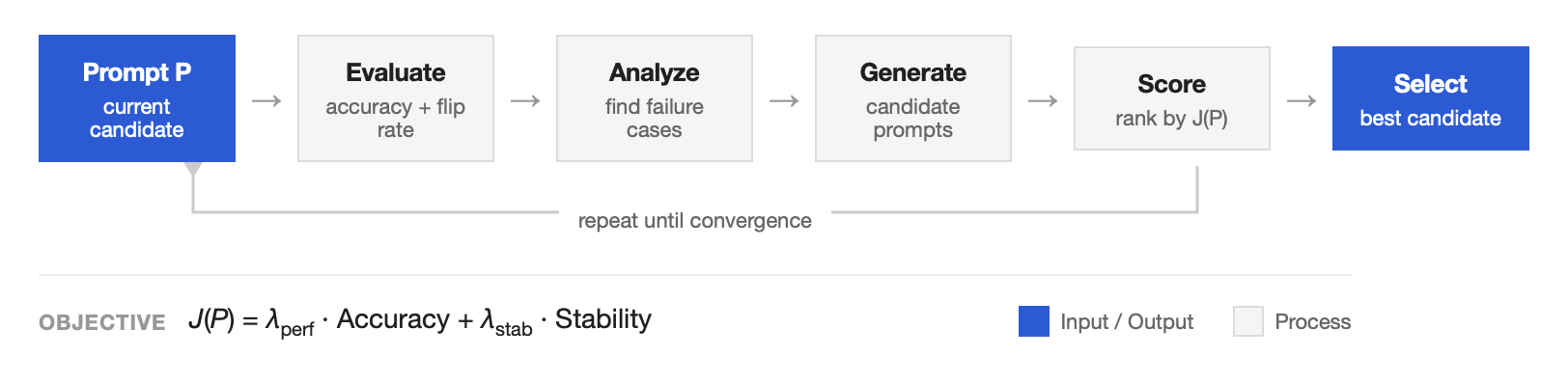}
\caption{Stability-aware prompt optimization loop. Starting from an initial prompt, we evaluate accuracy and flip rate (measured across paraphrased variants), identify failure cases, and use an LLM to generate candidate prompts conditioned on these failures. Candidates are scored on a joint objective balancing performance and stability, and the best-scoring prompt is selected for the next iteration.}
\label{fig:method}
\end{figure*}

Large language models (LLMs) are increasingly used for clinical abstraction over electronic health record (EHR) text, including phenotype classification, applicability judgements, and response correctness assessment. This work draws on our experience operating CHARM, a production clinical abstraction system deployed at Century Health for phenotype classification and chart review tasks. In deployed systems, however, the model is only one part of the stack. The same underlying model is wrapped in different prompts: vendor defaults, institution-specific templates, and local tweaks by teams trying to ``make it behave.'' In this setting, both predictive uncertainty and prompt sensitivity, i.e., how much model behaviour changes under semantically equivalent prompt edits, become central to safety and trust. A system that appears well calibrated under one prompt but becomes erratic under another is difficult to validate and harder to monitor once deployed.

Most work on medical LLMs has focused on accuracy and, increasingly, on calibration or abstention mechanisms, but largely treats the prompt as a fixed design choice. Existing studies characterise calibration and uncertainty on clinical QA or summarisation benchmarks, or explore selective prediction policies that abstain when confidence is low, yet they do so under a single prompt template and a single backend. As a result, we know relatively little about how uncertainty estimates behave when prompts change, or whether seemingly ``well calibrated'' models remain reliable once the surface form of the instruction is perturbed. This is problematic in realistic pipelines, where prompts differ between vendors, are iteratively tweaked by domain experts, or must be adapted to fit within product constraints.

In this paper we argue that, for clinical abstraction tasks where correctness is judged against unstructured evidence such as note excerpts or visit summaries, prompting and uncertainty should be treated as a joint problem rather than as separate concerns. Concretely, we study two grounded datasets and ask three questions: (i) how sensitive are different open and proprietary backends to prompt wording at the instance level on these tasks; (ii) how do standard uncertainty signals (probabilities, conformal sets) behave under prompt perturbations; and (iii) can we exploit sensitivity information to drive a dual-objective prompt optimisation loop that trades off task performance against prompt stability.

Our contributions are threefold.
\begin{itemize}
    \item We characterise prompt sensitivity across clinical abstraction tasks and multiple backends, measuring flip rates and calibration under prompt perturbations for both open-weight and proprietary models.

    \item We analyse the relationship between prompt stability and model uncertainty, finding that stable predictions tend to be more confident, though uncertainty alone does not fully predict sensitivity.

    \item We propose a dual-objective prompt optimisation loop that explicitly balances accuracy and stability, demonstrating that joint optimisation reduces flip rates across tasks and models.
\end{itemize}

\section{Related work}

\subsection{Prompt sensitivity and multi-prompt evaluation}

A growing body of work shows that LLM behaviour is highly sensitive to seemingly minor changes in prompt wording. Zhuo et al.\ introduce ProSA, a framework for prompt sensitivity analysis that operates at the instance level, defining PromptSensiScore to quantify disagreement across semantically equivalent prompts and relating this to decoding confidence, model size, and task difficulty \citep{zhuo2024prosa}. ProSA demonstrates that larger models are not uniformly more robust and that subjective, judge-based evaluations are particularly vulnerable to prompt perturbations on complex reasoning tasks.

At the task and benchmark level, Mizrahi et al.\ argue that single-prompt evaluation is fundamentally unreliable, and call for multi-prompt LLM evaluation \citep{mizrahi2024multiprompt}. They construct large sets of paraphrased instructions for existing benchmarks and show that both absolute scores and model rankings can change dramatically across templates. They propose metrics such as maximum performance, average performance and a combined score (CPS) to summarise behaviour over a prompt set and highlight that current ``state-of-the-art'' claims are highly prompt-dependent. Together, ProSA and multi-prompt evaluation provide complementary evidence that prompt formulation is a major, under-controlled source of variance, but they stop at measurement rather than proposing concrete optimisation procedures.

Other work analyses prompt sensitivity in more specialised settings (e.g.\ prompt engineering for natural language interfaces in software visualisation or multimodal models), but these typically either rely on small, domain-specific prompt sets or treat sensitivity qualitatively rather than building an optimisation objective around it \citep{atzberger2025sensitivity}. None of these studies connect prompt sensitivity to uncertainty estimation or selective prediction, and none use sensitivity signals to drive where to invest optimisation effort.

\subsection{Prompt optimisation and robustness}

A large literature targets automatic prompt optimisation for performance, usually on standard NLP tasks, under a fixed train/test distribution. Representative methods include RL-based optimisation of discrete prompts (e.g.\ RLPrompt) and search-based or LLM-in-the-loop rewriting approaches, which treat prompts as a policy to be tuned against a task-specific reward. These methods almost exclusively optimise a single metric such as accuracy, BLEU or task reward, and ignore stability across paraphrases or distributional shifts.

More recently, several lines of work explicitly study robust prompt optimisation. Li et al.\ formulate robust prompt optimisation under distribution shift, arguing that prompts tuned on a labelled source group often fail on an unlabelled target group; their Generalised Prompt Optimisation (GPO) framework incorporates unlabelled target data to improve worst-group performance while maintaining source performance \citep{li2023robust}. DRO-InstructZero treats instruction optimisation as a distributionally robust optimisation problem, maximising worst-case expected utility over an f-divergence ball around the evaluation distribution to obtain prompts that are more reliable under real-world uncertainty \citep{li2025dro}. Closely related work in safety focuses on robust optimisation of system prompts against jailbreaking or prompt injection attacks, where the robustness objective is defined in terms of adversarial success rather than semantic stability (e.g., BlackDAN: \citealp{wang2024blackdan}; AEGIS: \citealp{liu2025aegis}).

There is also an emerging strand of multi-objective prompt optimisation. SoS (``Survival of the Safest'') introduces an evolutionary framework that optimises prompts for both performance and safety \citep{sinha2024sos}, while ParetoPrompt formulates prompt optimisation as a multi-objective reinforcement-learning problem, exploring a Pareto front of prompts that trade off multiple metrics \citep{zhao2025paretoprompt}. These methods show that multi-objective formulations are useful in prompt optimisation, but they typically focus on performance vs safety or efficiency, not on semantic prompt stability.

Closest to our focus, Chen et al.\ introduce Prompt Stability Matters, which argues that automatic prompt generators for general-purpose multi-agent systems should account for prompt stability, defined as consistency of model responses across repeated executions for the same prompt \citep{chen2025promptstability}. They propose semantic stability as a criterion, train a LLaMA-based evaluator to score it, and build a stability-aware prompt generation system that uses stability feedback to improve both accuracy and output consistency. However, their notion of stability is primarily about stochastic reproducibility of a fixed prompt, rather than sensitivity to semantically equivalent prompt variants for the same input, and their experiments are conducted on general-purpose multi-agent workflows rather than clinical abstraction tasks.

Finally, error-driven prompt refinement methods such as ProTeGi and PO2G use feedback on model mistakes to update prompts, but still optimise a single performance metric without explicit stability terms \citep{pryzant2023protegi,lieander2025po2g}. We bridge these lines of work by using instance-level sensitivity signals to guide prompt optimisation on clinical abstraction tasks, connecting prompt stability to uncertainty estimation in a way that existing frameworks do not.

\section{Method}

\subsection{Problem setup and optimisation objective}

In deployed clinical systems such as CHARM, prompt optimisation typically involves iterative refinement with human feedback. Annotators review model outputs, provide corrections, and the prompt is updated to reduce errors on a labelled evaluation set. This process optimises for task accuracy but does not explicitly account for prompt stability---the extent to which predictions remain consistent under semantically equivalent prompt rephrasings.

We extend this approach by formulating prompt optimisation as a dual-objective problem. Let $F(P)$ denote the task performance (e.g., accuracy) of prompt $P$, and let $S(P)$ denote a stability measure (e.g., negative flip rate). We optimise:
\begin{equation}
J(P) = \lambda_{\text{perf}} \cdot F(P) + \lambda_{\text{stab}} \cdot S(P)
\end{equation}
where $\lambda_{\text{perf}}$ and $\lambda_{\text{stab}}$ control the trade-off between performance and stability. Setting $\lambda_{\text{stab}} = 0$ recovers accuracy-only optimisation; setting both weights equal gives balanced joint optimisation.

\subsection{Optimisation loop}

We implement an LLM-in-the-loop optimisation procedure that iteratively refines the prompt to improve the joint objective. Each iteration proceeds as follows.

\paragraph{Evaluation.} Given the current prompt $P$, we run inference on a held-out evaluation set to compute accuracy. To estimate stability, we generate $K$ paraphrased variants of $P$ using an LLM and measure the flip rate---the proportion of examples whose predictions change between the base prompt and its paraphrases.

\paragraph{Failure identification.} We identify two types of failure cases: (i) high-flip examples, where predictions are unstable across prompt variants, and (ii) misclassified examples, where the base prompt prediction is incorrect. These cases inform the next round of prompt generation.

\paragraph{Candidate generation.} An LLM proposes $N$ candidate prompts conditioned on the current prompt, its performance metrics, and concrete failure examples. This targeted feedback encourages edits that address specific error patterns rather than generic rephrasings.

\paragraph{Selection.} Each candidate is scored on the joint objective $J(P)$. If the best candidate improves over the current prompt, it is accepted; otherwise, the iteration makes no update. The loop terminates after a fixed number of iterations or when no improvement is found for several consecutive rounds.

In our experiments, we compare accuracy-only optimisation ($\lambda_{\text{stab}} = 0$) against joint optimisation ($\lambda_{\text{perf}} = \lambda_{\text{stab}} = 0.5$) to isolate the effect of the stability term.

\subsection{Uncertainty signals and metrics}

We extract uncertainty signals differently depending on model access. For local HuggingFace models, we compute class probabilities from per-token logprobs by taking a softmax over label scores, using the top-class probability as a confidence estimate. For Bedrock-hosted models, which provide only predicted labels, we rely on agreement across prompt variants: flip rate serves as a proxy for prediction uncertainty.

For models with probability access, we construct conformal prediction sets following standard split conformal inference. We use a nonconformity score of $1 - P(\text{label})$, set the coverage level to $\alpha = 0.1$, and partition the held-out set into calibration and evaluation subsets (50/50 split). We report empirical coverage and average set size.

We evaluate models along four dimensions: (i) \emph{performance}---accuracy and F1 by task, with log-loss and Brier score when probabilities are available; (ii) \emph{calibration}---reliability diagrams, expected calibration error (ECE), and maximum calibration error (MCE); (iii) \emph{sensitivity}---flip rate between base and variant prompts, and Jensen--Shannon divergence between prompt-induced distributions for models with probability access; and (iv) \emph{selective prediction}---coverage--accuracy curves using probability thresholds or conformal set size, abstaining when set size exceeds one or confidence falls below a threshold.

\subsection{Stability-Conditioned Uncertainty Analysis}
  \label{sec:stability-calibration}

To investigate whether prompt sensitivity manifests in model uncertainty estimates, we analyze the relationship between per-example flip rates and conformal prediction set sizes. This addresses a natural question: do examples that are unstable under prompt perturbation also exhibit higher predictive uncertainty?

\paragraph{Per-example flip indicator.}
For each example $x_i$ in the evaluation set, let $\hat{y}_i^{(0)}$ denote the prediction under the base prompt and $\hat{y}_i^{(k)}$ the prediction under the $k$-th paraphrased variant. We define the binary flip indicator and continuous flip rate as:
\begin{equation}
\text{flip}_i = \mathbf{1}\left[\exists k : \hat{y}_i^{(k)} \neq \hat{y}_i^{(0)}\right], \qquad \text{flip\_rate}_i = \frac{1}{K} \sum_{k=1}^{K} \mathbf{1}\left[\hat{y}_i^{(k)} \neq \hat{y}_i^{(0)}\right]
\end{equation}
where $K$ is the number of prompt variants.

\paragraph{Conformal set size.}
For models providing class probabilities, we construct conformal prediction sets using the base prompt predictions. Following standard split conformal inference~\cite{vovk2005algorithmic}, we partition the held-out set into calibration and evaluation subsets (50/50 split). For each evaluation example $x_i$, the conformal set $\mathcal{C}_i$ contains all labels $y$ such that the nonconformity score $1 - P(y \mid x_i)$ falls below the calibrated threshold at level $\alpha = 0.1$. We record the set size $|\mathcal{C}_i|$ and whether the true label is covered, $y_i \in \mathcal{C}_i$.

\paragraph{Stratified analysis.}
We partition examples into \emph{stable} ($\text{flip}_i = 0$) and \emph{unstable} ($\text{flip}_i = 1$) groups and report: (i)~mean conformal set size per group, (ii)~empirical coverage per group, and (iii)~Spearman rank correlation between $\text{flip\_rate}_i$ and $|\mathcal{C}_i|$.

A positive correlation would indicate that model uncertainty, as captured by conformal set size, reflects prompt sensitivity, suggesting that standard uncertainty quantification partially accounts for instability. Conversely, low or no correlation would imply that calibration and prompt stability measure distinct phenomena, reinforcing the need to optimize for both objectives independently.

This analysis is restricted to models with access to output probabilities (local HuggingFace models). For API-based models providing only predicted labels, we report group-level flip rates without conformal conditioning.

\section{Data}

We study prompt sensitivity and uncertainty on two chart-grounded clinical abstraction settings: MedAlign applicability and correctness, and an internal multiple sclerosis (MS) subtype abstraction task derived from the internal note corpus. In both cases, models must reason over unstructured clinical text (visit notes, summaries) to produce labels that reflect clinician judgement rather than surface form alone.

\subsection{MedAlign}

MedAlign is a clinician-generated benchmark for instruction following on electronic medical records (EHRs) \citep{fleming2024medalign}. The dataset contains 983 natural-language instructions written by 15 practising clinicians across 7 specialties, paired with 276 longitudinal EHR timelines. For 303 instruction-EHR pairs, clinicians provided reference responses and ranked evaluations of six large language model outputs. MedAlign is released as a test-only benchmark under a research data use agreement and is designed to capture realistic information-seeking and documentation needs rather than exam-style question answering.

We work with MedAlign v1.3, which includes two additional annotation tables that we use to define our tasks. First, an applicability table provides an is\_applicable field indicating whether a given instruction is actually supported by the information present in the chart. Second, a clinician-reviewed responses table contains binary\_correct labels indicating whether a particular model response is judged correct or incorrect given the same instruction and chart context.

From these tables we derive two binary classification tasks:

\textbf{MedAlign applicability:} ``Is the instruction supported by the chart?'', with labels \{Yes, No\} taken directly from is\_applicable.

\textbf{MedAlign correctness:} ``Does the model response align with clinician judgement for this instruction and chart?'', with labels \{Correct, Incorrect\} mapped from binary\_correct $\in \{0,1\}$.

For each record we assemble a textual input consisting of the instruction, the chart evidence, the clinician reference summary when available, and the model response (for correctness). Chart evidence is prefixed with a ``Chart evidence:'' header and separated from the instruction and other sections by blank lines. When a response is missing we insert the literal string ``<empty response>'' so that true blanks can be distinguished from parsing errors. All text is decoded as UTF-8 and we preserve original casing and punctuation. We drop rows with invalid or missing labels and deduplicate records by instruction identifier, model name and annotator, where applicable.

Because MedAlign is explicitly test-only, we do not use it for any parameter tuning or prompt pre-training. Instead, we cap the evaluation subset at 100 examples for applicability and 200 for correctness. For each task we draw a single subset of labelled rows and use the same subset and random 50/50 calibration--evaluation split across all models and prompt variants. This design keeps comparisons fair while keeping computational cost and manual inspection effort manageable.

\subsection{MS subtype}

Our second dataset is an internal EHR-derived corpus focusing on MS subtype abstraction from neurology notes. Clinically, MS is commonly described in terms of course phenotypes such as relapsing--remitting MS (RRMS), primary progressive MS (PPMS) and secondary progressive MS (SPMS), with clinically isolated syndrome and ``not documented'' or ``unspecified'' categories also used in practice \citep{lublin2014defining,klineova2018clinical}. In routine care, this information is often encoded inconsistently across structured fields and free-text documentation, which makes it a natural target for LLM-based abstraction.

The MS subtype dataset is constructed from de-identified notes for patients with a confirmed MS diagnosis. Each note, or visit-level bundle of notes, is paired with a target label reflecting the documented subtype at that time. For this study we formulate a six-way classification problem with the label set: Clinically Isolated Syndrome (CIS), Relapsing-Remitting MS (RRMS), Primary Progressive MS (PPMS), Secondary Progressive MS (SPMS), Not Documented, and Unknown. Labels are derived from a combination of structured phenotype modifiers and manual review by domain experts. We do not use imaging or laboratory data, and we do not attempt to infer latent phenotype from longitudinal trajectories; the task is strictly to abstract what is documented in the note.

All internal data notes are passed through an institutional de-identification pipeline that removes direct identifiers (names, dates of birth, contact details) and obvious quasi-identifiers. We retain section headings and paragraph breaks but do not perform additional clinical concept normalisation. As with MedAlign, we treat the dataset as a held-out evaluation corpus for this study. To keep experiments comparable in scale, we cap the internal data note evaluation subset at 100 notes per run. We again use a single held-out set shared across all models and prompt variants, and apply a random 50/50 calibration--evaluation split for conformal prediction.

\subsection{Task construction and edge cases}

Across both datasets we focus on tasks where correctness is judged with respect to unstructured evidence, not purely against pre-existing structured labels. In MedAlign applicability, the ground-truth label reflects whether the requested operation is supported by the chart as presented. Some instructions are inherently non-applicable, for example, tooling prompts or future protocol requests that cannot be resolved from the current EHR snapshot, even though their phrasing appears reasonable. These cases introduce a degree of annotation noise and highlight that applicability judgements depend on chart coverage, not only on linguistic plausibility.

In MedAlign correctness, labels reflect clinician assessments of particular model outputs. Responses may be partially correct, over-confident or hallucinated; the binary label encodes clinical acceptability rather than exact string match to a reference. This makes the task distribution different from generic question answering and closer to real-world audit of model behaviour.

In the MS subtype, the ``Not documented'' class is intended to capture notes where no explicit subtype is present. In practice, some notes contain ambiguous or implicit descriptions of disease course. We retain these borderline examples but flag them for sensitivity analyses, as they are precisely the cases where over-interpretation by a model could be harmful.

For all tasks we retain basic metadata such as anonymised patient or visit identifiers, instruction identifiers and model names (for MedAlign correctness). This enables stratified analyses by visit, instruction type or source model, and supports future work on temporal drift and per-patient aggregation.

\subsection{Ethical and privacy considerations}

Both tasks involve sensitive clinical information. MedAlign is de-identified and provided under a research data use agreement by its original authors, who explicitly position it as a test-only benchmark \citep{fleming2024medalign}. We follow this guidance and do not attempt to reconstruct or augment the underlying EHR data. The MS subtype corpus is derived from an internal EHR system and is de-identified under the host institution's governance processes. Access is restricted under an internal data-use protocol that prohibits re-identification attempts and external sharing of raw notes. All experiments in this work are observational and retrospective; we do not generate patient-facing outputs, and we use the datasets solely to study model behaviour, uncertainty and prompt sensitivity.

\section{Experiments and results}

\subsection{E1: Stability and Model Certainty}

To test whether prompt stability relates to model certainty, we analyze the relationship between per-example flip rates and prediction margin (the difference between top-two class probabilities). For each example, we compute the flip rate across $K=3$ prompt variants and the margin from the base prompt's output distribution.

\begin{figure}[t]
      \centering
      \includegraphics[width=\columnwidth]{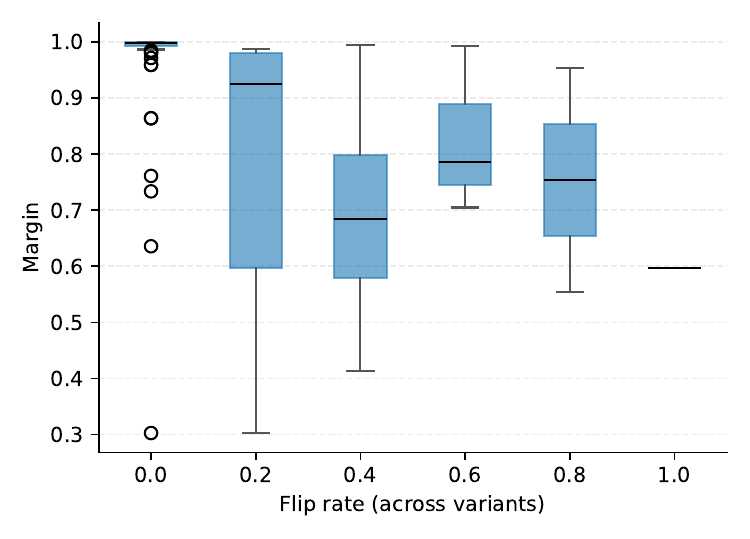}
      \caption{Prediction margin versus flip rate for MedGemma 4B on MedAlign applicability. Stable examples (flip rate = 0) have high margins, indicating confident predictions. As flip rate increases, margins decrease and become more variable, suggesting that uncertain predictions are more prone to flipping under prompt variation.}
      \label{fig:e1-margin}
  \end{figure}

Figure~\ref{fig:e1-margin} shows results for MedGemma 4B on MedAlign applicability as box plots grouped by flip rate. Examples that never flip (flip rate = 0) have margins concentrated near 1.0, indicating high confidence. As flip rate increases, median margin decreases and variability grows. This pattern supports the intuition that stable predictions are more trustworthy: when the model is certain, it is also robust to prompt variation.

% However, the relationship is not absolute. Some stable examples have low margins, and some high-flip examples retain moderate confidence. This suggests that while certainty is a useful signal for stability, it cannot fully substitute for explicit stability measurement. We observe similar patterns across other HuggingFace models. (Appendix, Figure~\ref{fig:e1-multimodel}).

\subsection{E2: Accuracy Does Not Guarantee Stability}

A natural question is whether improving accuracy automatically improves prompt stability. To test this, we generate multiple semantically equivalent prompt variants for a single task and model, evaluate each on accuracy and flip rate, and examine whether higher-accuracy prompts are also more stable.

\begin{figure}[htbp]
    \centering
    \includegraphics[width=\columnwidth]{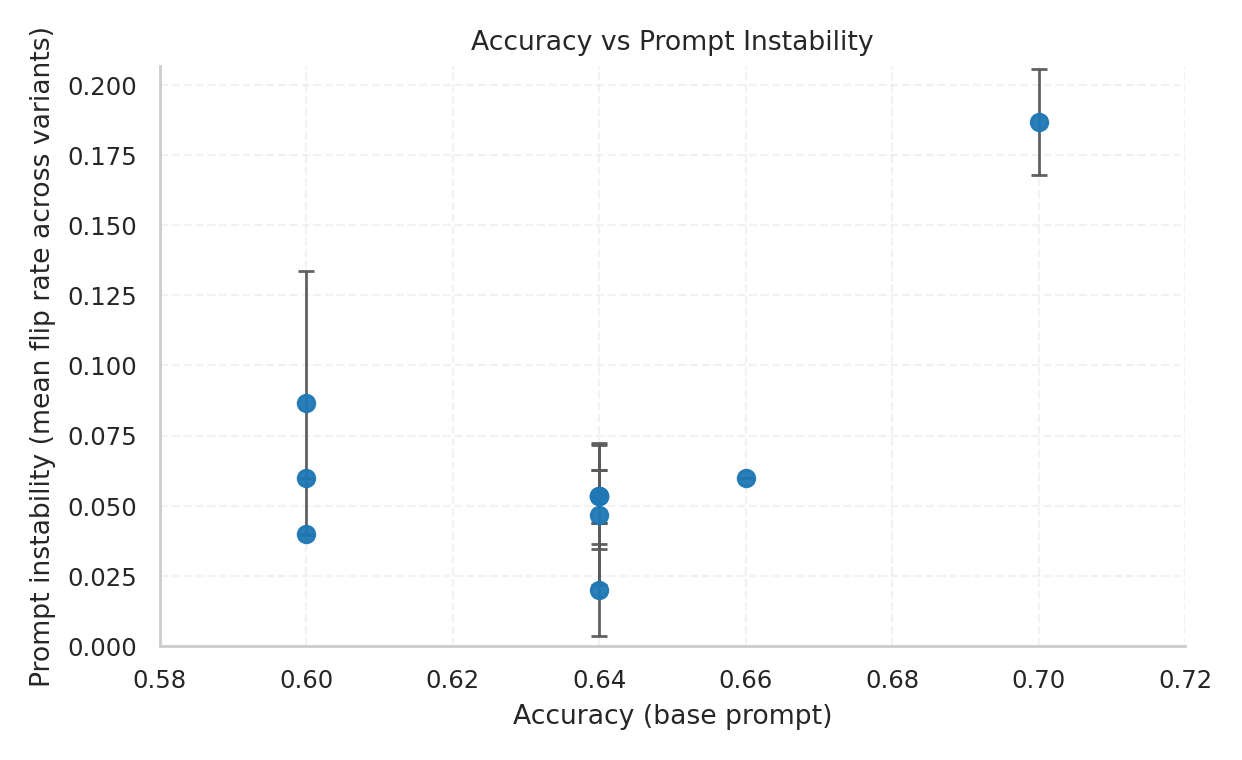}
    \caption{Accuracy vs flip rate for MedAlign applicability using Llama 3 70B. Each point is one candidate prompt; the y-axis is the mean flip rate across three paraphrased variants. The vertical spread at similar accuracy levels illustrates that improving accuracy alone does not guarantee prompt stability.}
    \label{fig:e2-stability}
\end{figure}

Figure~\ref{fig:e2-stability} shows results for Llama 3 70B on MedAlign applicability. Each point represents a candidate prompt; the x-axis shows accuracy and the y-axis shows mean flip rate across $K=3$ paraphrased variants. If accuracy implied stability, we would expect a negative correlation with points concentrated in the upper-left (low accuracy, high flip) and lower-right (high accuracy, low flip) regions.

Instead, we observe substantial vertical spread: prompts with similar accuracy can have very different flip rates. Some high-accuracy prompts are unstable (upper-right region), while some lower-accuracy prompts are relatively stable. This decoupling suggests that optimizing for accuracy alone does not guarantee prompt robustness, motivating explicit stability objectives. We observe similar patterns across tasks and models.
% (Appendix, Figure~\ref{fig:e2-multimodel}).

\subsection{E3: Joint Optimization Improves Stability}

\begin{figure*}[htb]
      \centering
      \includegraphics[width=\textwidth]{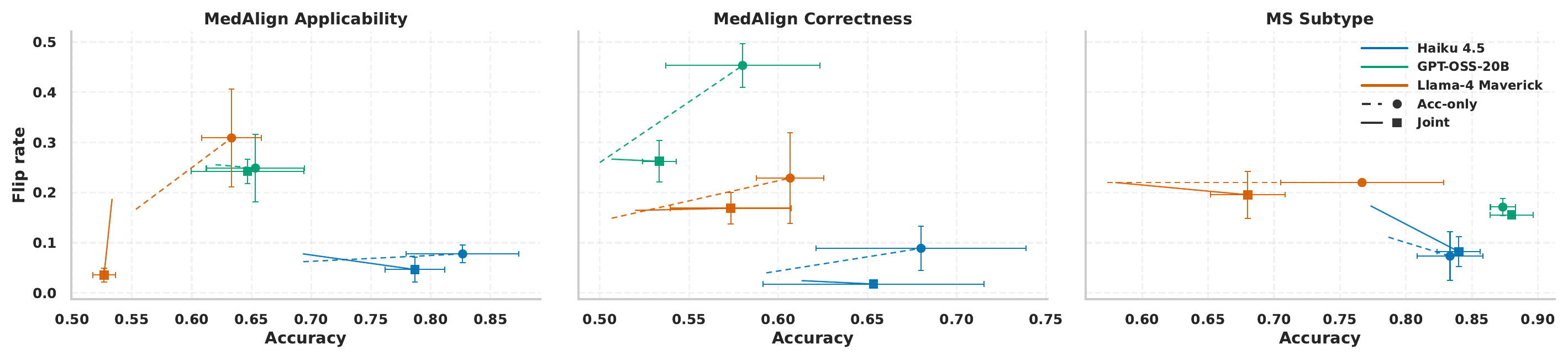}
      \caption{Summary of E3 optimizer sweeps across tasks and models. Each segment shows the mean start$\rightarrow$end trajectory (3 seeds) in accuracy--flip space; endpoints include error bars. Color encodes model; line style/marker encodes objective (accuracy-only vs joint). Adding a stability term systematically shifts endpoints toward lower flip rates, often with modest accuracy trade-offs.}
      \label{fig:e3-summary}
  \end{figure*}

We run the stability-aware prompt optimizer across three clinical tasks (MedAlign applicability, MedAlign correctness, and MS subtype) using three Bedrock models (Claude Haiku 4.5, Llama-4 Maverick, GPT-OSS-20B). Each setting is repeated with three random seeds on a fixed stratified subset ($N=50$). We compare accuracy-only optimization ($\lambda_{\text{perf}}=1,\ \lambda_{\text{stab}}=0$) against joint optimization ($\lambda_{\text{perf}}=\lambda_{\text{stab}}=0.5$). Table~\ref{tab:e3_sweep} reports mean $\pm$ std over seeds for end-of-run accuracy and flip rate. Figure~\ref{fig:e3-summary} visualizes the optimization trajectories across all tasks and models.

Across the nine model-task combinations, joint optimization improves stability in eight cases. The largest gains occur when baseline stability is poor: for Llama-4 Maverick on MedAlign applicability, flip rate drops from 0.309 to 0.036, and for GPT-OSS-20B on MedAlign correctness, from 0.453 to 0.262. The accuracy trade-off varies: in some cases (GPT-OSS-20B on MS subtype) accuracy slightly improves under joint optimization, while in others (Llama-4 Maverick on applicability) it decreases. The one exception is Haiku on MS subtype, where baseline stability is already high and joint optimization shows no additional benefit.

\begin{table}[t]
\centering
\small
\begin{tabular}{llcc}
\toprule
Task / Model & Setting & Acc. (end) & Flip (end) \\
\midrule
\multicolumn{4}{l}{\textbf{MedAlign Applicability}} \\
Haiku 4.5 & Acc-only & 0.827 $\pm$ 0.047 & 0.078 $\pm$ 0.017 \\
Haiku 4.5 & Joint & 0.787 $\pm$ 0.025 & 0.047 $\pm$ 0.025 \\
Llama-4 Maverick & Acc-only & 0.633 $\pm$ 0.025 & 0.309 $\pm$ 0.098 \\
Llama-4 Maverick & Joint & 0.527 $\pm$ 0.009 & 0.036 $\pm$ 0.014 \\
GPT-OSS-20B & Acc-only & 0.653 $\pm$ 0.041 & 0.249 $\pm$ 0.067 \\
GPT-OSS-20B & Joint & 0.647 $\pm$ 0.047 & 0.242 $\pm$ 0.025 \\
\midrule
\multicolumn{4}{l}{\textbf{MedAlign Correctness}} \\
Haiku 4.5 & Acc-only & 0.680 $\pm$ 0.059 & 0.089 $\pm$ 0.044 \\
Haiku 4.5 & Joint & 0.653 $\pm$ 0.062 & 0.018 $\pm$ 0.006 \\
Llama-4 Maverick & Acc-only & 0.607 $\pm$ 0.019 & 0.229 $\pm$ 0.090 \\
Llama-4 Maverick & Joint & 0.573 $\pm$ 0.034 & 0.169 $\pm$ 0.031 \\
GPT-OSS-20B & Acc-only & 0.580 $\pm$ 0.043 & 0.453 $\pm$ 0.043 \\
GPT-OSS-20B & Joint & 0.533 $\pm$ 0.009 & 0.262 $\pm$ 0.041 \\
\midrule
\multicolumn{4}{l}{\textbf{MS Subtype}} \\
Haiku 4.5 & Acc-only & 0.833 $\pm$ 0.025 & 0.073 $\pm$ 0.048 \\
Haiku 4.5 & Joint & 0.840 $\pm$ 0.016 & 0.082 $\pm$ 0.030 \\
Llama-4 Maverick & Acc-only & 0.767 $\pm$ 0.062 & 0.220 $\pm$ 0.000 \\
Llama-4 Maverick & Joint & 0.680 $\pm$ 0.028 & 0.196 $\pm$ 0.047 \\
GPT-OSS-20B & Acc-only & 0.873 $\pm$ 0.009 & 0.171 $\pm$ 0.017 \\
GPT-OSS-20B & Joint & 0.880 $\pm$ 0.016 & 0.156 $\pm$ 0.003 \\
\bottomrule
\end{tabular}
\caption{E3 sweep results: accuracy-only vs joint optimization (mean $\pm$ std over 3 seeds). Joint optimization consistently reduces flip rate, with variable accuracy trade-offs depending on model and task.}
\label{tab:e3_sweep}
\end{table}

\section{Discussion}

Our results support three main findings about prompt stability in clinical LLM systems, which we summarize below and then discuss in terms of their practical implications.

\paragraph{Accuracy and stability are decoupled at the prompt level.}
E2 demonstrates that within a single model, prompts with similar accuracy can have very different flip rates. This is distinct from the model-level trend (Appendix, Figure~\ref{fig:app-fliprate}) where larger models tend to be both more accurate and more stable. The implication is that selecting a capable model does not guarantee prompt-level robustness; prompt engineering must explicitly account for stability.

\paragraph{Stability correlates with model certainty.}
E1 shows that stable examples, those whose predictions do not flip under prompt paraphrases, tend to have higher prediction margins (Spearman $\rho = 0.204$ between margin and stability). This supports the intuition that confident predictions are more trustworthy: when the model is certain, it is also more robust to prompt variation. However, the correlation is modest, indicating that certainty alone is not a reliable proxy for stability.

\paragraph{Stability is optimizable.}
E3 demonstrates that adding an explicit stability term to the optimization objective improves stability in 8 of 9 model-task combinations, sometimes with modest accuracy trade-offs. This suggests that prompt stability should be treated as a first-class objective rather than assumed to follow from accuracy improvements.

\paragraph{Bridging prompt sensitivity and prompt optimization.}
Prior work on prompt sensitivity~\citep{zhuo2024prosa,mizrahi2024multiprompt} establishes that LLM behavior varies substantially across paraphrased instructions, but stops at measurement. Separately, prompt optimization methods~\citep{pryzant2023protegi,li2023robust} focus on improving accuracy without accounting for stability. Our work bridges these literatures by using sensitivity signals to drive optimization, showing that stability can be explicitly targeted rather than merely observed.

\paragraph{A complementary signal for clinical AI validation.}
Current best practices for validating clinical LLM systems emphasize accuracy, calibration, and selective prediction~\citep{fleming2024medalign}. Our findings suggest prompt stability should be added to this checklist. A system may appear well-calibrated under its development prompt yet behave erratically when that prompt is paraphrased by a downstream team. Stability testing provides a complementary validation signal that existing metrics do not capture.

\paragraph{Implications for multi-team deployment.}
In realistic clinical deployments, the same underlying model is often wrapped in different prompts by different stakeholders: vendor defaults, institution-specific templates, and local modifications by clinical informaticists. Our results suggest this practice carries hidden risk. A prompt validated by one team may become unstable when another team makes seemingly innocuous edits. Explicitly optimizing for stability, or at minimum measuring it, could reduce unexpected failures when prompts drift across teams or over time.

\paragraph{Stability as a trust signal.}
The correlation between stability and model certainty (E1) suggests a practical heuristic: predictions that are robust to prompt variation may be more trustworthy than those that flip. This aligns with intuitions from ensemble methods, where agreement across models signals reliability. Here, agreement across prompts serves an analogous role. While the correlation is modest, it provides initial evidence that stability could inform selective prediction policies, abstaining on examples that are both uncertain and prompt-sensitive.

\paragraph{Integration into production.}
These findings have informed ongoing development of the CHARM system at Century Health, where stability metrics are now tracked alongside accuracy during prompt validation. When prompts are updated or new models are evaluated, we measure flip rates across paraphrased variants as part of the release process, flagging prompts that show high instability for additional review before deployment.

  \subsection{Limitations}

  \paragraph{Scale.}
  Our experiments use capped evaluation sets (50--200 examples per task) and only $k=3$ prompt paraphrases for flip rate estimation. The stability-calibration correlation ($\rho = 0.204$, $p = 0.056$) is borderline significant, and larger samples may reveal stronger or weaker relationships. The E3 sweeps use $N=50$ examples with 3 seeds, which limits statistical power for detecting small effects.

  \paragraph{Paraphrase generation.}
  Flip rates depend on the paraphrases used. We generate variants via LLM rewriting, which may not reflect the full range of prompt variations encountered in deployment (e.g., vendor templates, institution-specific formatting, non-native speaker phrasings). The stability we measure is relative to our paraphrase distribution, not absolute.

  \paragraph{Stability-calibration analysis restricted to HF models.}
  The E1 analysis joining flip rates with conformal set sizes is only possible for models providing output probabilities (HuggingFace). For Bedrock models (Claude, Llama, Qwen), we can measure flip rates but cannot assess the stability-calibration relationship. This limits generalizability to API-based deployments.

  \paragraph{Single optimization method.}
  We evaluate one LLM-in-the-loop optimizer architecture. We did not compare against alternative prompt optimization methods (evolutionary search, gradient-based, random search), so we cannot claim our specific approach is optimal, only that incorporating a stability term into \emph{some} optimization objective is beneficial.

  \paragraph{No downstream clinical validation.}
  We show that joint optimization reduces flip rates, but we do not demonstrate that this translates to improved clinical outcomes or reduced harm. The link between prompt stability and patient safety remains indirect.

  \paragraph{Task scope.}
  Our experiments cover two clinical abstraction settings (MedAlign and MS subtype). Generalization to other clinical NLP tasks, such as entity extraction, temporal reasoning, or multi-document summarization, remains untested.

\bibliographystyle{plainnat}
\bibliography{refs}

\newpage
\appendix

\section*{Appendix}
\subsection*{PSS Validation}
\label{app:pss}

\begin{figure}[htb]
\centering
\includegraphics[width=0.6\columnwidth]{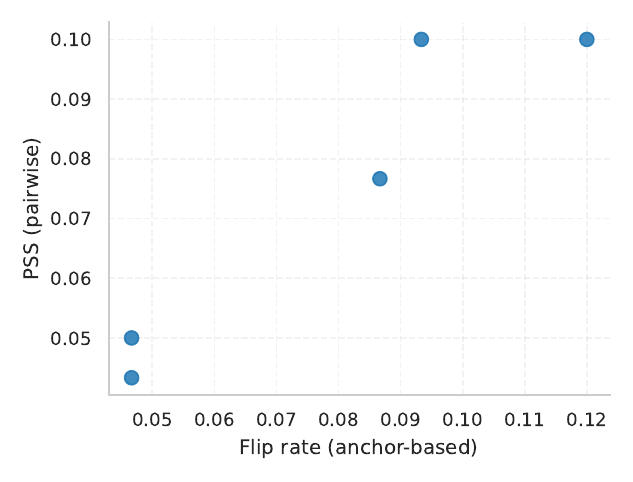}
\caption{Flip-rate vs.\ ProSA-style PSS on a 50-example subset (5 prompts, $k=3$ paraphrases). The strong correlation validates our anchor-based flip-rate as a practical proxy for the symmetric PSS metric.}
\label{fig:pss_vs_flip}
\end{figure}

To validate our anchor-based flip-rate metric against the published ProSA framework, we compute the ProSA-style instance-level prompt sensitivity score (PSS) on a small subset using a base prompt plus $k=3$ paraphrases. Figure~\ref{fig:pss_vs_flip} shows a strong correlation between flip-rate and PSS (5 prompts, MedAlign applicability with Haiku), indicating our simpler anchor-based metric tracks the symmetric PSS measure.

\subsection*{Optimizer Trajectories}
\label{app:trajectories}

\begin{figure*}[htb]
\centering
\includegraphics[width=\textwidth]{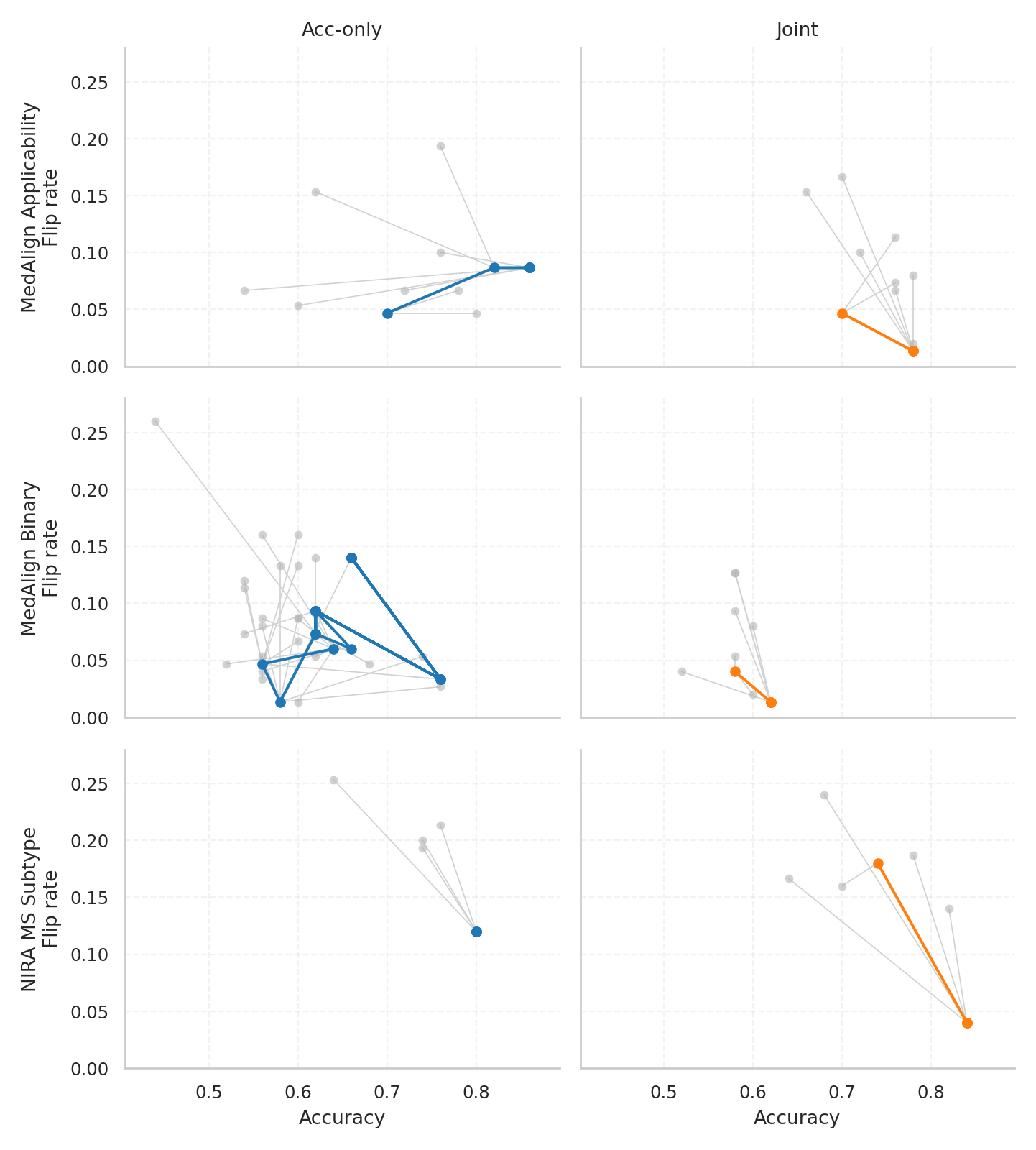}
\caption{Optimizer trajectories for Claude Haiku 4.5 (seed 1). Each panel shows accepted prompts (solid line) and candidate proposals (faint grey) in accuracy--flip-rate space. Left column: accuracy-only optimization; right column: joint optimization. Rows: MedAlign applicability (top), MedAlign binary (middle), MS subtype (bottom).}
\label{fig:trajectories}
\end{figure*}

Figure~\ref{fig:trajectories} shows representative optimizer trajectories in accuracy--flip-rate space, comparing accuracy-only and joint optimization settings across tasks. Each panel traces the sequence of accepted prompts, while faint grey candidate points and connectors illustrate the local search space explored at each iteration. Rows correspond to tasks and columns to the optimization setting (accuracy-only vs.\ joint accuracy-stability). These trajectories visualize how the optimization dynamics differ under the two objectives: accuracy-only runs tend to move horizontally (improving accuracy without regard for stability), while joint optimization traces paths that also push downward in flip-rate.

\subsection*{Model-Level Sensitivity and Calibration}
\label{app:model-comparison}

\begin{figure}[htb]
\centering
\includegraphics[width=\columnwidth]{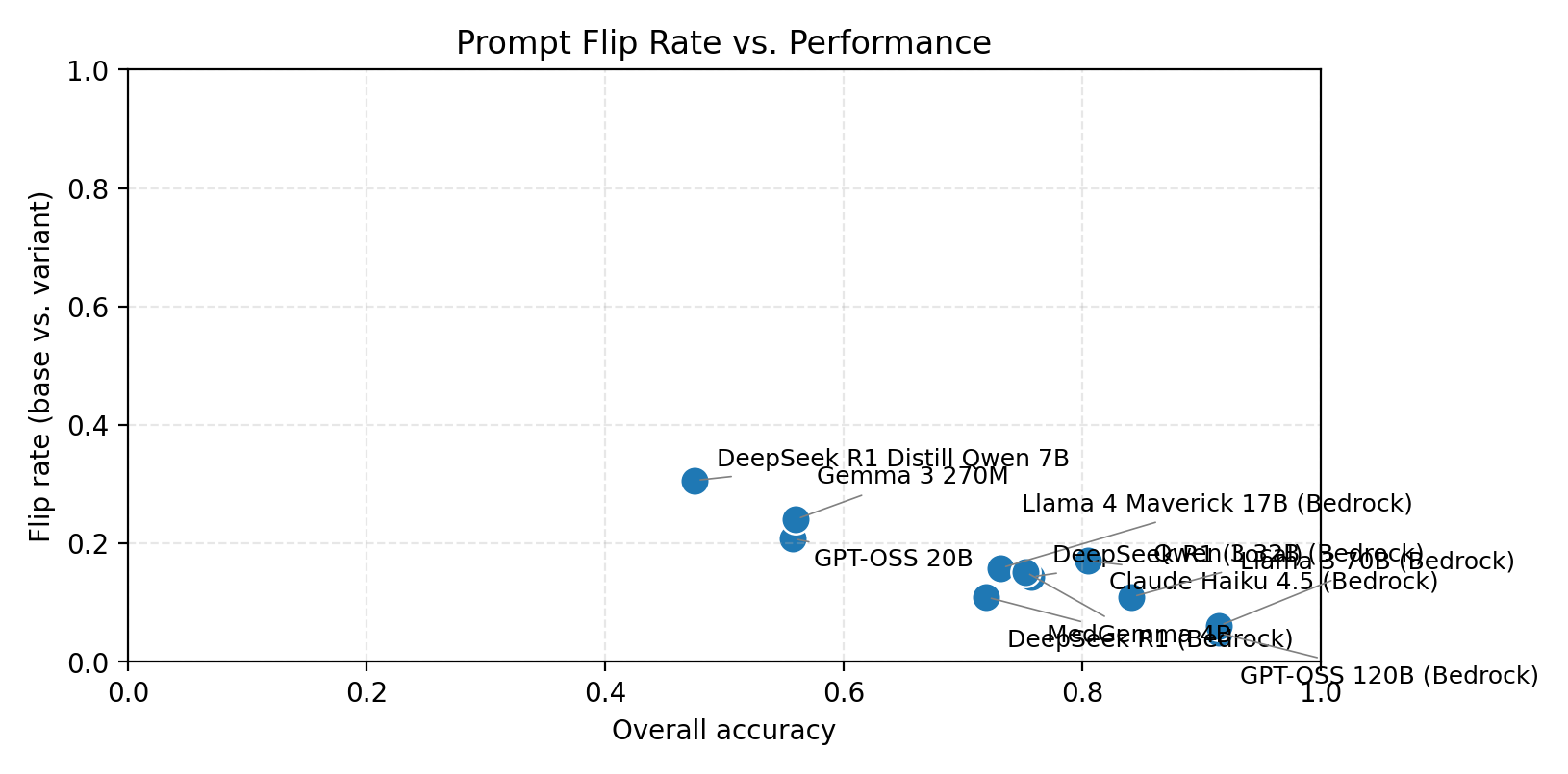}
\caption{Flip rate vs accuracy trade-off across models on MedAlign applicability. Larger models tend to be both more accurate and more stable at the model level, though this does not guarantee prompt-level stability.}
\label{fig:app-fliprate}
\end{figure}

While our main experiments focus on the relationship between stability and uncertainty at the example level, we also examined model-level patterns. Figure~\ref{fig:app-fliprate} shows flip rate versus accuracy across models on MedAlign applicability. Larger models tend to occupy the high-accuracy, low-flip region, while smaller models cluster in the low-accuracy, high-flip corner. This model-level trend suggests that capacity improvements benefit both metrics, but does not guarantee prompt-level stability within a given model (see E2).

\begin{figure}[htb]
\centering
\includegraphics[width=0.75\columnwidth]{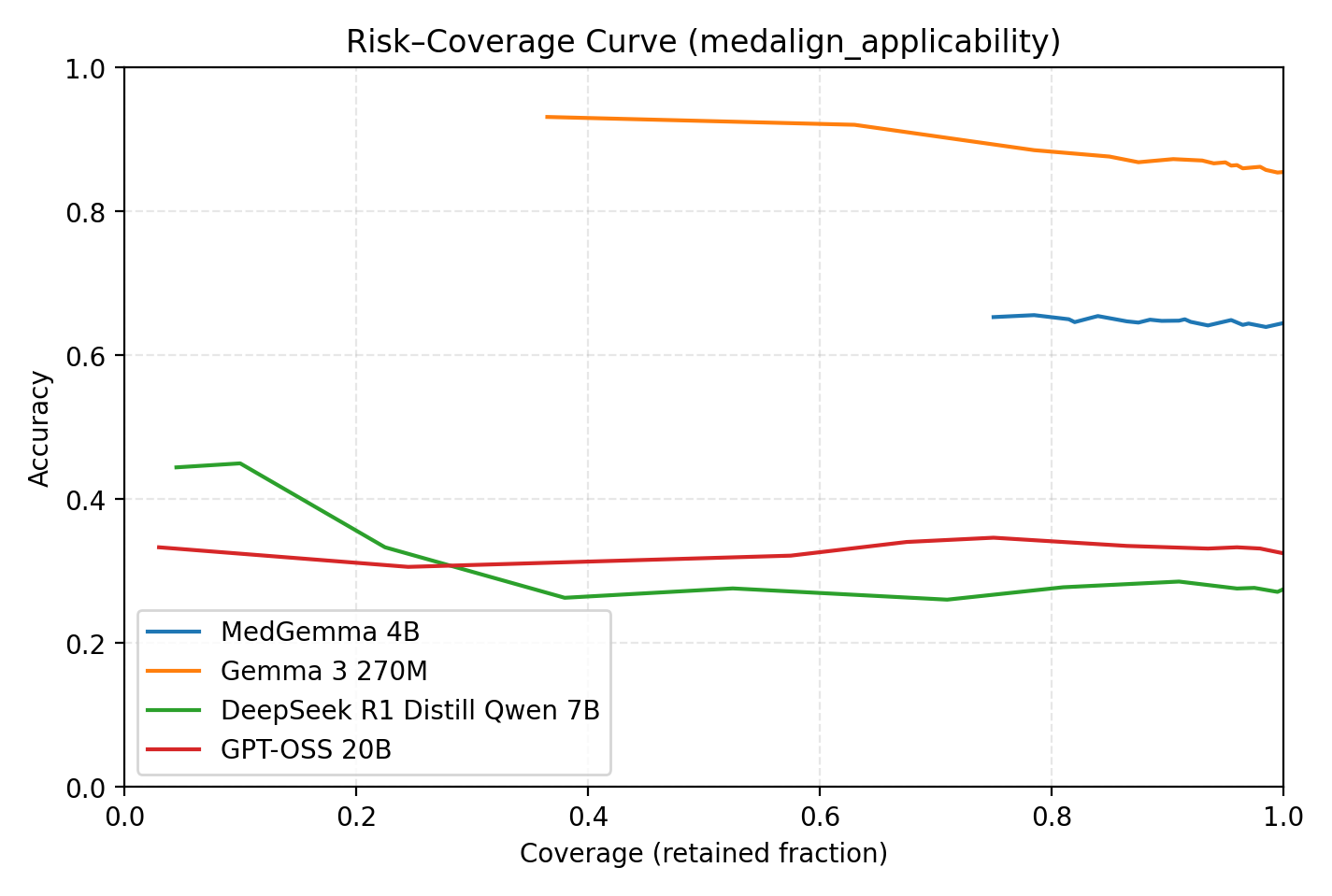}
\caption{Conformal prediction results on MedAlign applicability across HuggingFace models. Coverage targets can be met with very different abstention costs and retained accuracy.}
\label{fig:app-conformal}
\end{figure}

Figure~\ref{fig:app-conformal} shows conformal prediction coverage and set size trade-offs across HuggingFace models. Models achieve similar coverage targets ($\sim$0.90--0.97) with varying abstention costs (set sizes), illustrating that calibration quality differs substantially across backends.

\end{document}